\documentclass{article}

\usepackage{arxiv}

\usepackage[utf8]{inputenc} 
\usepackage[T1]{fontenc}    
\usepackage{hyperref}       
\usepackage{url}            
\usepackage{booktabs}       
\usepackage{amsfonts}       
\usepackage{nicefrac}       
\usepackage{microtype}      
\usepackage{lipsum}
\usepackage{graphicx,subcaption}
\usepackage{amsmath}
\usepackage{amssymb}

\newcommand{\bW}{\mathbf{W}}
\newcommand{\bx}{\mathbf{x}}
\newcommand{\blambda}{\mathbf{\lambda}}

\def\bZ{\boldsymbol{Z}}
\def\bu{\boldsymbol{u}}
\def\bX{\boldsymbol{X}}
\def\bY{\boldsymbol{Y}}

\def\bw{\boldsymbol{w}}
\def\ba{\boldsymbol{r}}
\def\beps{\pmb{\varepsilon}}
\def\bSig{\pmb{\Sigma}}

\def\bx{\boldsymbol{x}}

\def\bPi{\pmb{\Pi}}

\title{Autoregressive Hidden Markov Models with partial knowledge on latent space applied to aero-engines prognostics}
\date{}

\author{
  Pablo Juesas,  Emmanuel Ramasso$^{*}$, S\'ebastien Drujont and Vincent Placet \\
  Department of Applied Mechanics\\
  Institut FEMTO-ST,  UBFC/UFC/ENSMM/CNRS/UTBM, \\
  24 Rue Alain Savary, 25000 Besan\c con, France\\
  $^{*}$\texttt{emmanuel.ramasso@femto-st.fr}  
}

\begin{document}
\maketitle

\begin{abstract}

[This paper was initially published in PHME conference in 2016, selected for further publication in International Journal of Prognostics and Health Management.]\\

This paper describes an Autoregressive Partially-hidden Markov model (ARPHMM) for fault detection and prognostics of equipments based on sensors' data.  It is a particular dynamic Bayesian network that allows to represent the dynamics of a system by means of a Hidden Markov Model (HMM) and an autoregressive (AR) process. The Markov chain assumes that the system is switching back and forth between internal states while the AR process ensures a temporal coherence on sensor measurements. A sound learning procedure of standard ARHMM based on maximum likelihood allows to iteratively estimate all parameters simultaneously. This paper suggests a modification of the learning procedure considering that one may have prior knowledge about the structure which becomes partially hidden. 
The integration of the prior is based on the Theory of Weighted Distributions which is compatible with the Expectation-Maximization algorithm in the sense that the convergence properties are still satisfied. We show how to apply this model to estimate the remaining useful life based on health indicators. 
The autoregressive parameters can indeed be used for prediction while the latent structure
can be used to get information about the degradation level. The interest of the proposed method for 
prognostics and health assessment is demonstrated on CMAPSS datasets. 

\end{abstract}


\section{AR-Markov modelling for prognostics}

Autoregressive (AR) models have been shown to be appropriate for time-series modelling in various fields
such as econometric \cite{ng2002forecasting} and climate forecasting \cite{arclimate}. In condition monitoring, 
it is generally used for fault detection by establishing an AR model using healthy conditions under various loads
and using this model on the observed data recorded in in-service conditions in order to trend the residual signals.
Such approaches does not make use of analytical description of faults or collection of typical fault 
patterns \cite{FarrarBook,Serdio13}. 
Such a method has been used for gear monitoring by \cite{wang2002autoregressive}, and \cite{yan04} used a similar approach together with a logit model to determine the probability of failure of an elevator door motion system.
\cite{Thanagasundram2008975} suggested to use a pole representation of an AR process to detect bearings faults.
\cite{saha09} compared ARIMA (AR with integrated moving average) with two other models for battery prognostics.

In structural safety, AR models were used by \cite{he1997system,huang2001structural} 
to identify structural dynamic characteristics of system subjected to ambient excitations. 
\cite{Ling2011868} 
suggested to use an ARMA to characterize and reconstruct fatigue loads for prognostics application 
on mechanical components. The authors proposed to adapt the parameters of this model by Bayesian updating to 
accommodate variability in loading and data sparsity. 

\cite{Lehman15} were interested in the modelling of switching 
autoregressive dynamics from multivariate vital sign time series in order to stratify mortality risks of intensive 
care units patients receiving particular treatments. In this biomedical application, 
the authors made use of a combination of an AR process and a Hidden Markov Model (HMM) called ARHMM. 
Such a model is able to cope with multistate non-stationary systems which are generally encountered in 
PHM applications. 

ARHMM were also used for wind turbine monitoring by \cite{Ailliot201292} where the authors were 
interested in the statistical representation of wind time series. They made use of a 
hidden Markov chain which represents the weather types and allows to switch between several autoregressive
models that describe the time evolution of the wind speed. Such switching models are particular 
well suited for PHM applications to represent the dynamical behavior of complex systems \cite{Serir2013213,Lim16}. 

Initially proposed for speech recognition \cite{rabiner1989tutorial}, an ARHMM is a particular 
dynamic Bayesian network that draws benefits of an AR model and HMM. 
A standard and sound learning procedure of this model based on maximum likelihood (ML) allows parameters 
to be estimated iteratively and simultaneously. 
This paper suggests to use ARHMM for fault detection and prognostics of equipments based on sensors' data. 
A modification of the learning procedure is also proposed to enable one to add prior knowledge on the latent structure.
This modification allows, in some way, to decrease the attachment of the model to the data that can be observed in practice 
in various probabilistic models learned with the ML approach. 

Following previous work on the integration of 
prior on latent variables for fault detection \cite{come2009learning,ramasso2009contribution,Cherfi2012,ramasso2014making},
the objective is to enable the users of probabilistic models to represent two kinds of knowledge \cite{Dubois07,denoeux2013maximum}:
\begin{itemize}
\item Generic knowledge about the data generating process and corresponding to random uncertainty.
It pertains to a population of observables such as historical facts, laws of physics, statistical and 
common sense knowledge.
\item Specific knowledge, also called relative knowledge or factual evidence, is about a given 
realization of the data. It pertains to a particular situation and is related to a domain or a discipline.
\end{itemize}
Specific knowledge is of key importance for PHM applications. 
It is not necessarily related to statistics and is generally partial because the 
observation process is imperfect due to lack of knowledge. It generally aims at improving skills 
of a method trained by generic knowledge.

The integration of the prior suggested in this paper for ARHMM 
is based on the Theory of Weighted Distributions (TWD) \cite{Patil2002} which 
is compatible with the Expectation-Maximization (EM) algorithm in the sense that the convergence properties are 
still satisfied. It makes use of concepts initially developed by \cite{come2009learning,denoeux2013maximum}
based on Dempster-Shafer's theory of belief functions and of \cite{Juesas16} using the TWD to include prior in EM-based 
learning procedures. 

The resulting model is called Autoregressive Partially-Hidden Markov Model (ARPHMM) and is described in the next 
section. It is then shown to be well suited for remaining useful life (RUL) based on health indicators
with an illlustration on CMAPSS datasets \cite{frederick2007user,saxena2008damage}.

\section{Markov switching model with soft prior: General formulation}

\begin{figure}[ht]
\centering
\includegraphics[width=0.99\columnwidth]{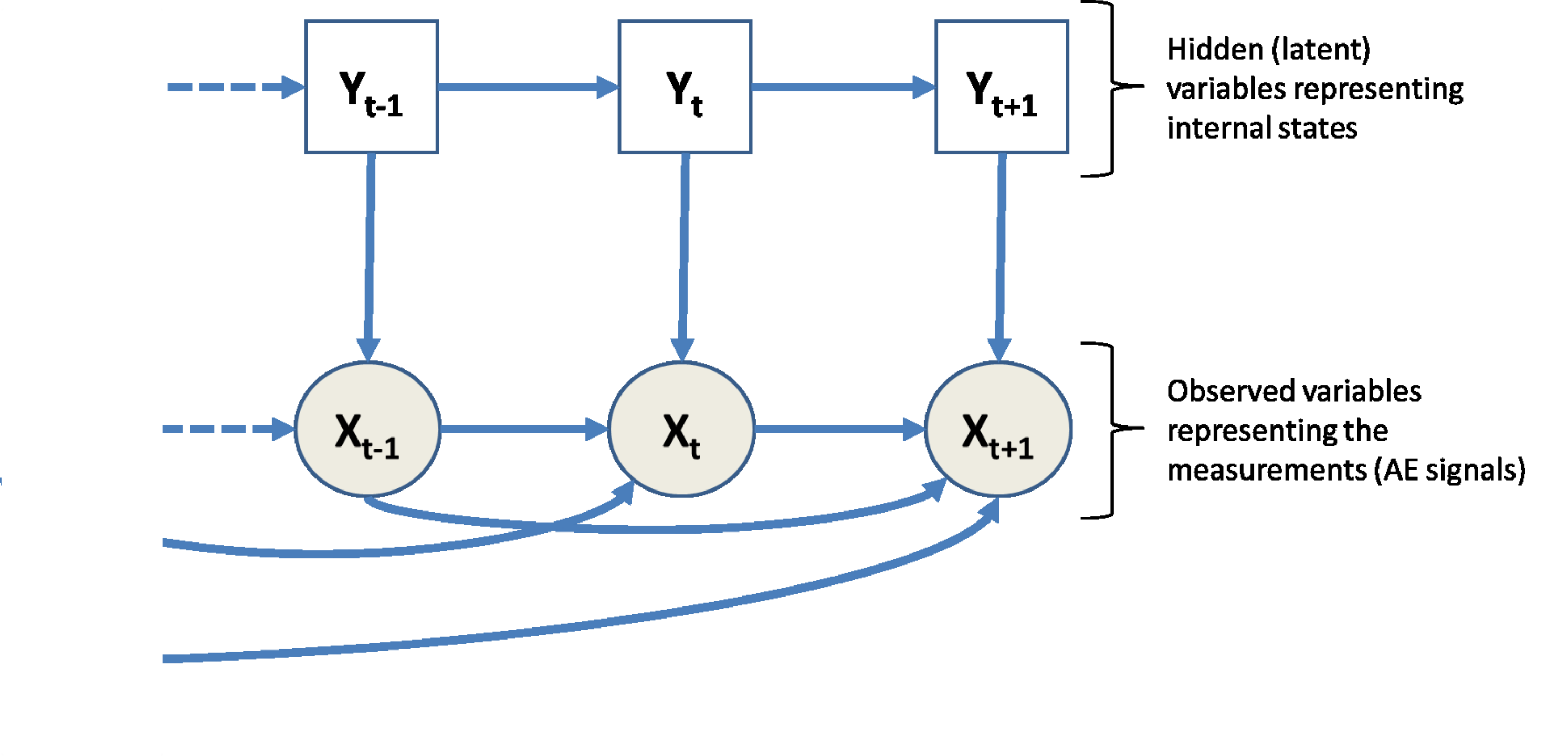}
\caption{Graphical model of an ARHMM: Rounded boxes $\bX_t$ represent continuous observed variables (measurements such as AE signals at 
time $t$), rectangular-shaped boxes $\bY_t$ represent hidden discrete variables. The AR process is represented by the 
links between measurements. \label{fig:arphmm}}
\end{figure}

A measurement at time $t$, $\bx_t$ is mathematically represented as a weighted sum of the 
previous measurements plus an error term, where the weights are defined conditionally to each state:
\begin{align}
\bx_{t} = -\sum_{\delta=1}^{\Delta}\ba_{\delta}(y_{t})\bx_{t-\delta} + \beps_{t}(y_{t}),\:\: 1\leq t \leq T
\label{eq:xtar}
\end{align} 
The noise term $$\beps_{t}(y_{t}) \sim N(0, \bSig_{y_t})$$ is assumed to be a Gaussian 
with zero mean and a covariance matrix $\bSig_{y_t}$ automatically adjusted for each 
hidden state given the data in the learning phase. 
The AR coefficients for the $i$-th state are denoted as $\ba_{\delta}(y_{t}=i)$ where $\delta=1\dots \Delta$ 
is the time lag. The set of AR coefficients is given by:
\begin{equation}
 \mathbf{B}_i =\Big(\ba_{1}(i), \dots, \ba_{\delta}(i), \dots, \ba_{\Delta}(i)\Big)\\
\end{equation}
The switching between internal states is governed by a stochastic process taking the form of a Markov chain
and depicted in Figure~\ref{fig:arphmm}. It is represented by a transition matrix $\textbf{A}$ 
with elements $a_{ij}=p(y_t=j|y_{t-1}=i)$ (probability of going into state $j$ at time $t$ 
given the state was $i$ at $t-1$).
The prior probability of the chain is denoted as $\bPi = [\pi_1\dots \pi_K]$, where $\pi_i$ is the probability to 
be in state $i$ at time $t=1$.
 
\subsection{Incorporating prior knowledge on latent variables}

The problem is to estimate the parameters
\begin{align}
\blambda=(\textbf{A},\boldsymbol{\pi},\mathbf{B}_i,{\bSig}_i),\:\: 1\leq i \leq K
\label{eq:paramARP}
\end{align}
in presence of uncertain and imprecise prior information about the hidden variables. 


The prior is supposed to take the form of distributions over possible internal states given AE signals and represent users' 
beliefs before some evidence is taken into account. For practical use, we assume the prior to be
uncertain and imprecise so that we can cover different learning paradigms: 
Unsupervised learning (health states $\bY$ are supposed hidden), 
supervised learning (states corresponds to class fully known), 
semi-supervised learning (combination of both previous cases) and
partially-supervised learning were some health states can be known and accompanied by a confidence 
 degree $\bW = [\bw_1 ;\dots; \bw_t ;\dots ;\bw_T]$, with 
 $\bw_t = [w_t(1), \dots, w_t(i), \dots, w_t(K)]$ and $w_t(i) \geq 0$. This is the most general case: 
 \begin{itemize}
  \item When $w_t(i)=1$ for a given state $i$ and $w_t(j)=0, j\neq i$ then the supervised case is recovered; 
  \item When $\forall t, \forall i, w_t(i)=0$, then the unsupervised case is recovered.  
 \end{itemize}

In order to estimate the parameters of an ARPHMM in a sound manner when some prior knowledge $\bW$ about the hidden 
states is available, we suggest an approach using the TWD described 
by Patil \cite{Patil2002} and we derive the optimal solution in terms of maximum likelihood. 
It follows a similar reasoning to \cite{Juesas16,Ramasso14} and has connections with \cite{denoeux2013maximum}.

\subsection{Inference and learning in ARPHMM}
\label{model}




The parameters (Eq.~\ref{eq:paramARP}) are optimized by an Expectation-Maximization (EM)
learning procedure \cite{dempster77}. In the E-step, we evaluate the expectation of 
the hidden variables given the data; In the M-step, the auxiliary function $Q$ has to be maximized
in order to ensure that the likelihood will increase at each iteration, where $Q$ (at iteration $q$) given by:
\begin{equation}
 \begin{array}{rcl}
Q(\blambda,\blambda^{(q)}) &=& \displaystyle\mathbb{E}_{\blambda^{(q)}}[\log(L(\blambda;\bZ)|\bX] \\
 &=& \displaystyle\sum_{\bY} p(\bY|\bX,\blambda^{(q)}) \log L(\blambda;\bZ)  
 \end{array}
 \label{eq:Q2A}
\end{equation}
with $\bZ = (\bX, \bY)$. This expression requires to express the complete-data likelihood function which, for 
an ARPHMM, is given by:
\begin{equation}
\begin{array}{rcl}
L(\blambda;\bZ) &=& \displaystyle p(y_{1};\bPi)\bigg(\prod_{t=2}^{T}p(y_{t}|y_{t-1};\mathbf{A})\bigg) \times \\
& & \displaystyle\prod_{t=1}^{T}p(\bx_{t}|y_{t};\ba_\Delta(y_t),\bSig_{(y_t)}) 
\end{array}
\label{eq:lik}
\end{equation}
Using the Theory of Weighted Distributions (WDT) and for any positive weights $\mathbf{W}$, 
Eq.~\ref{eq:Q2A} can be modified as follows:
\begin{equation}
 \begin{array}{rcl}
Q(\blambda,\blambda^{(q)}) &=& \displaystyle\mathbb{E}_{\blambda^{(q)}}[\log L(\blambda;\mathbf{z})|\mathbf{x}, \mathbf{w}] \\
    &=& \displaystyle\frac{\sum_{\mathbf{y}} w(\mathbf{y}) p(\mathbf{y}|\mathbf{x},\blambda^{(q)}) 
    \log L(\blambda;\mathbf{z})}{\mathbb{E}_{\blambda}[w(\mathbf{y})]} \label{eq:Q2B}
 \end{array}
\end{equation}
This adaptation of EM for the ARHMM allows one to easily incorporate prior beliefs about the hidden states 
in a sound manner since EM still converges thanks to the normalisation.

We can expand the expression of $Q$, derive the expression with respect to the parameters to get the 
parameters at iteration $q+1$ of the modified EM. For the Markov chain, we have:
\begin{subequations}
 \begin{align}
\pi_{i}^{(q+1)}&=\gamma_{1i}^{(q)}\\
a_{ij}^{(q+1)}&=\frac{\sum_{t=2}^{T}\xi_{t-1,t,i,j}^{(q)}}{\sum_{t=2}^{T}\sum_{l=1}^{K}\xi_{t-1,t,i,l}^{(q)}} ,
\end{align}
\end{subequations}
We can show that the expression of the posterior probabilities $\gamma$ and $\xi$ can be obtained similarly to 
\cite{ramasso2014making}, using a modified forward-backward algorithm.

For the observation model, the noise covariance is given by:
\begin{equation}
 \begin{array}{rcl}
\displaystyle\bSig_{i}^{(q+1)}&=&\displaystyle\frac{1}{\sum_{t=1}^{T}\gamma_{ti}^{(q)}}\sum_{t=1}^{T}\gamma_{ti}^{(q)}\Big[\bx_{t}+
\sum_{\delta=1}^{\Delta}\ba_{\delta}^{(q)}(i) \\ 
& & \displaystyle\bx_{t-\delta}\Big] \Big[\bx_{t}+\sum_{\delta=1}^{\Delta}\ba_{\delta}^{(q)}(i)\bx_{t-\delta}\Big]^T ,
 \end{array}
\end{equation}
and the expression of the AR coefficients defined as:
\begin{equation}
 \begin{array}{rcl}
\mathbf{B}_i^{(q+1)}&=&\Big(\ba_{1}^{(q+1)}(i), \dots, \ba_{\delta}^{(q+1)}(i), \dots, \ba_{\Delta}^{(q+1)}(i)\Big)\\
&=&-\displaystyle\Big[\sum_{t=1}^{T}\gamma_{ti}^{(q)}\bx_{t}\bu_{t-1}^{T}\Big]\Big[\sum_{t=1}^{T}\gamma_{ti}^{(q)}\bu_{t-1}\bu_{t-1}^{T}\Big]
 \end{array}
\end{equation}
with
\begin{equation}
\bu_{t-1}=\Big(\bx_{t-1},\bx_{t-2},...,\bx_{t-\Delta}\Big)^{T} ,
\end{equation}
where the likelihood $b_{i}(\bx_{t})$ given the hidden state $i$ is given by:
\begin{equation}
b_{i}(\bx_{t}) = \mathcal{N}(\bx_{t} + \sum_{\delta=1}^{\Delta} \ba_{\delta}(i) \bx_{t-\delta} \;|\; 0,\bSig_{i}) 
\end{equation}

The forward pass is useful to evaluate the likelihood of the model. It is also of particular interest since it
may be defined with respect to the prior on the latent structure:
\begin{subequations}
 \begin{align}
\alpha_{1i}^{(q)}&=\pi_{i}^{(q)} \;w_{1i} \;b_{i}(\bx_{1}),\\
\alpha_{t,j}^{(q)}&=b_{j}(\bx_{t})\; \underbrace{w_{tj}}_{prior} \;\sum_{i}\alpha_{t-1,i}^{(q)}\; a_{ij}^{(q)} \label{eq:alpha}
\end{align}
\end{subequations}
and the likelihood of the observed data given the model is computed as:
\begin{equation}
L(\blambda^{(q)};\bX,\bW) =\sum_{i=1}^{K}\alpha_{Ti}
\end{equation}

\subsection{ARPHMM for health assessment and prognostics}
\label{sect:arphmmprop}

Health assessment can be performed by inferring the hidden state at the current time $t$. As in standard HMM, 
it is made possible by either a forward or a forward-backward passes or applying the Viterbi algorithm. 
The specificity of the proposed approach is to be able to exploit prior on the latent structure.

The remaining useful life can be estimated by a direct propagation computed, for instance, 
by a similarity-based approach. 
The principle is to look for a training instance in the historical data that is similar to the currently 
observed data and to consider that the latter would evolve in the same way \cite{wangieee}. The computation of the similarity
is however of key importance \cite{wangieee,Ramasso14}. 

Likelihood-based approaches does not always generalize well 
to unobserved (quite different) cases. This could be a limitation for health assessment 
and prognostics on real systems. 
Bayesian approaches have been used for tackling such problems in PHM, mostly during the prediction phase or for RUL 
estimation (for instance to update noise characteristics) and specifically to integrate prior on models' parameters
\cite{saha2008uncertainty,sankararaman2015uncertainty}. 

The Theory of Weighted Distributions used in this paper plays a similar role to the Bayesian approach but
specifically on the latent variables. Its interest holds particularly in the possibility to be used in MLE-based learning 
which represents a learning paradigm that is widely used in PHM-related publications. 
The use of prior on latent variables' configuration allows to condition some areas of the feature space which is 
expected to make those ML-based models more specific. 

Practically, the ARPHMM can be used for health assessment and prognostics as follows:
\begin{itemize}
\item Learning for prognostics: Build an ARPHMM by considering the RUL as output in the AR process. 
Consider various parameterizations of the prior ($w$) if the internal states have no meaning. 
The learning procedure thus estimates the mapping between the RUL and the data conditioned on the prior. 
 \item Health assessment on testing data: 
 Apply the forward-backward propagations and find the most probable state, or the Viterbi algorithm. 
 If no prior is available during inference, then $w_{t,i}=1$. If some prior information are available, then it can be used
 in the propagations (Eq.~\ref{eq:alpha}). 
 \item RUL estimation: For the testing phase, find the likeliest model by applying the forward pass together with 
 either the prior used for training (by assuming that the initial wear of both training and testing data are similar)
 or an external prior if available. Then deduce the RUL by merging the closest instances.
\end{itemize}
Note that RUL estimation could also be performed by sampling the underlying state-space model (Eq.~\ref{eq:xtar})
which is not studied in this paper and let for future work. 

\section{Illustration on turbofan datasets}

\subsection{Datasets description}

The turbofan datasets were generated using the CMAPSS simulation environment
that represents an engine model of the 90,000 lb thrust class \cite{frederick2007user,saxena2008damage}. 
The authors used a number of editable input parameters to specify operational profile, closed-loop controllers, 
environmental conditions. Some efficiency
parameters were modified to simulate various degradations in different sections of the engine system. 
Selected fault injection parameters were varied to simulate continuous degradation trends. 

The datasets generated possess unique characteristics that make them very useful and suitable for developing 
classification and prognostics algorithms: 
Multi-dimensional response from a complex non-linear system, 
high levels of noise, effects of faults and operational conditions, and 
plenty of units simulated with high variability. Benchmarking of prognostics algorithms on those datasets 
has been proposed and discussed in~\cite{RamassoSaxena2014}. 

\begin{figure}[ht]
 \centering 
{\includegraphics[width=0.95\columnwidth]{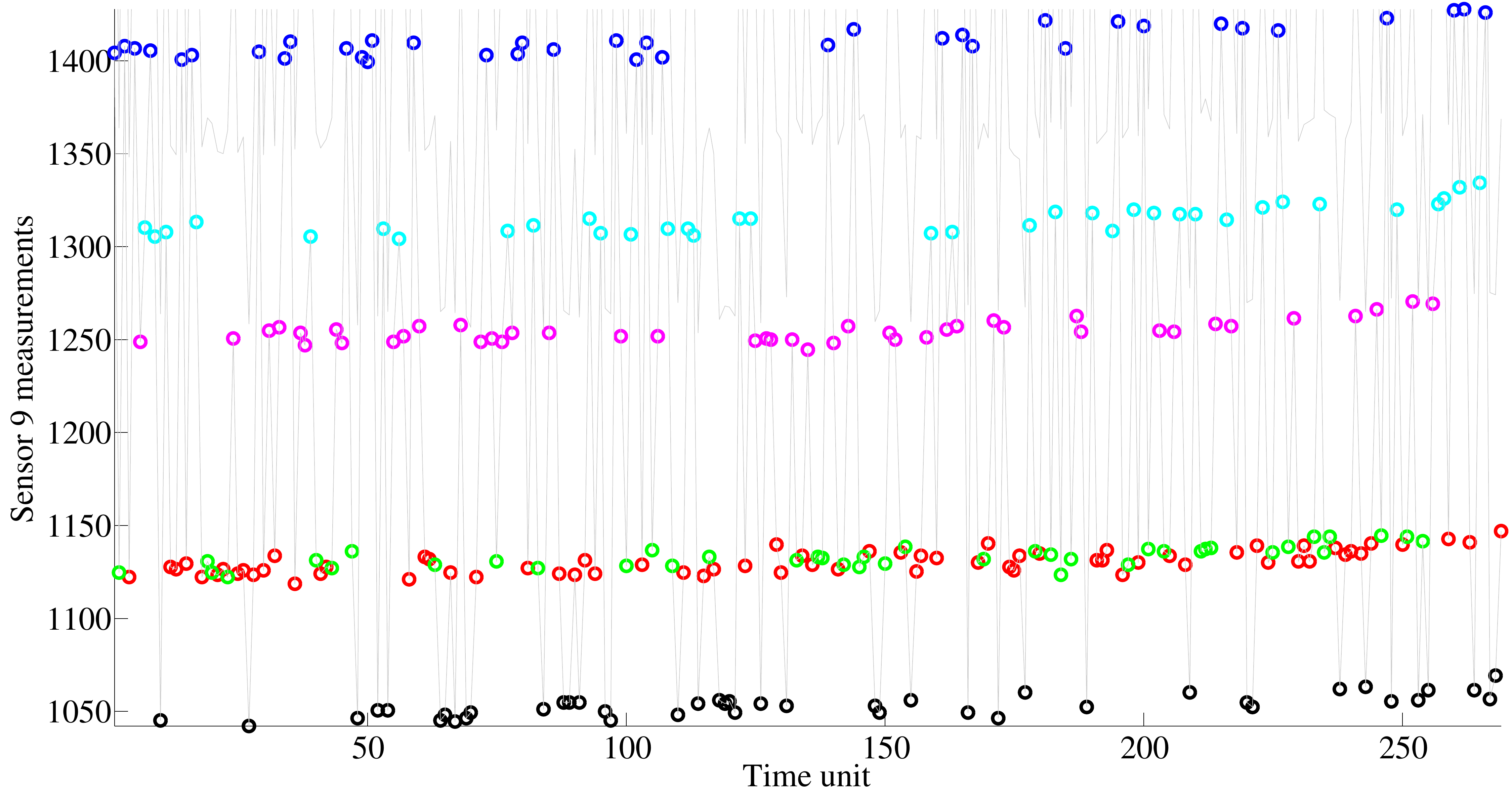}} 
\caption{Operating conditions in each regime: Sensor measurements
are locally linear . \label{fig:ds2b}}
\end{figure}

Figure~\ref{fig:ds2b} (taken from \cite{Ramasso14}) depicts one of the sensor measurements from a 
healthy situation to failure, as well as the evolution of those values in each of the 
six operating conditions (for instance landing, take-off, cruse and so on). 

\begin{figure*}
        \centering
        \begin{subfigure}[b]{0.45\textwidth}
                \includegraphics[width=0.95\textwidth]{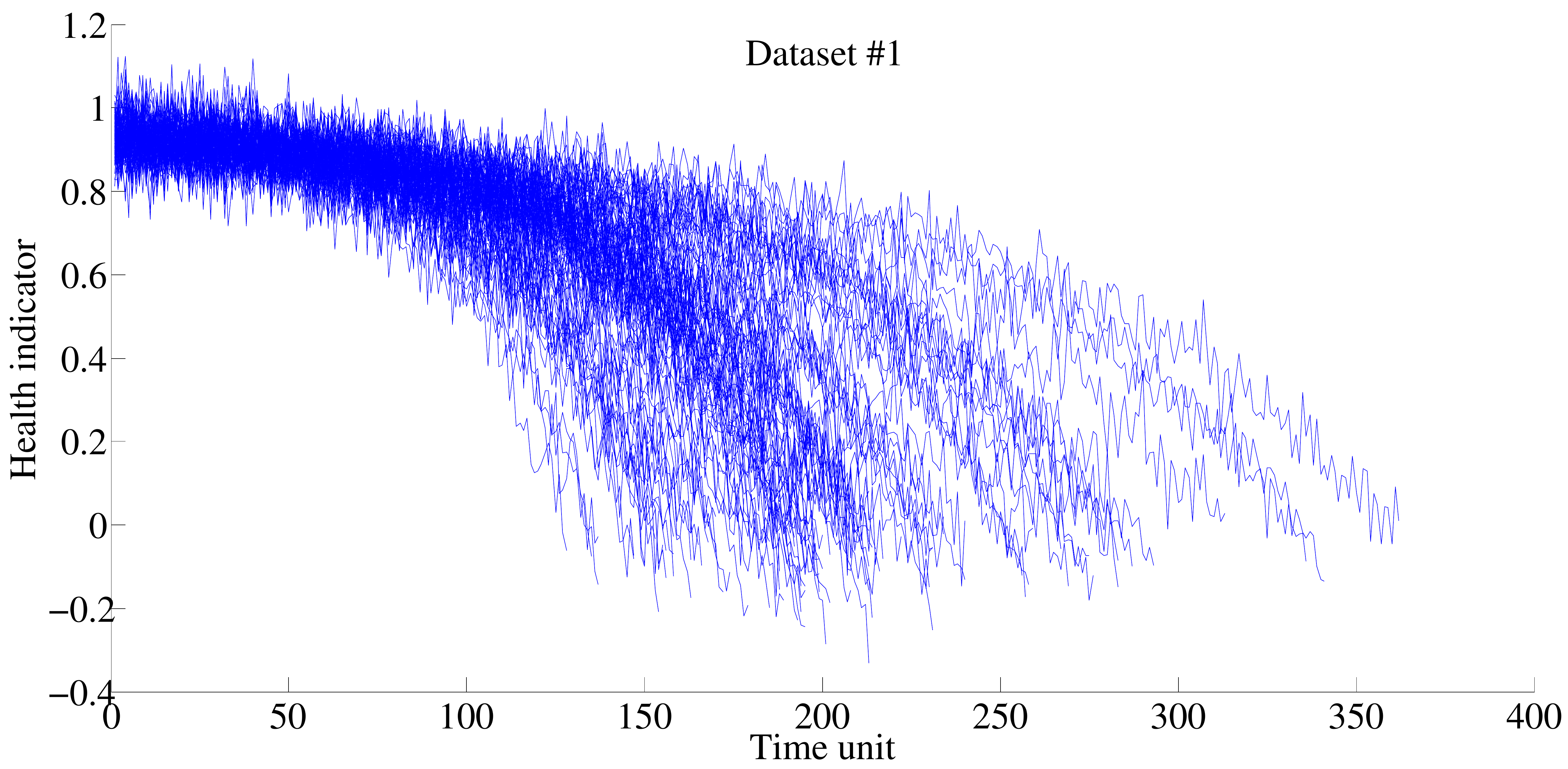}
                \caption{\#1\label{fig:hida1}}
        \end{subfigure}
        \centering
        \begin{subfigure}[b]{0.45\textwidth}
                \includegraphics[width=0.95\textwidth]{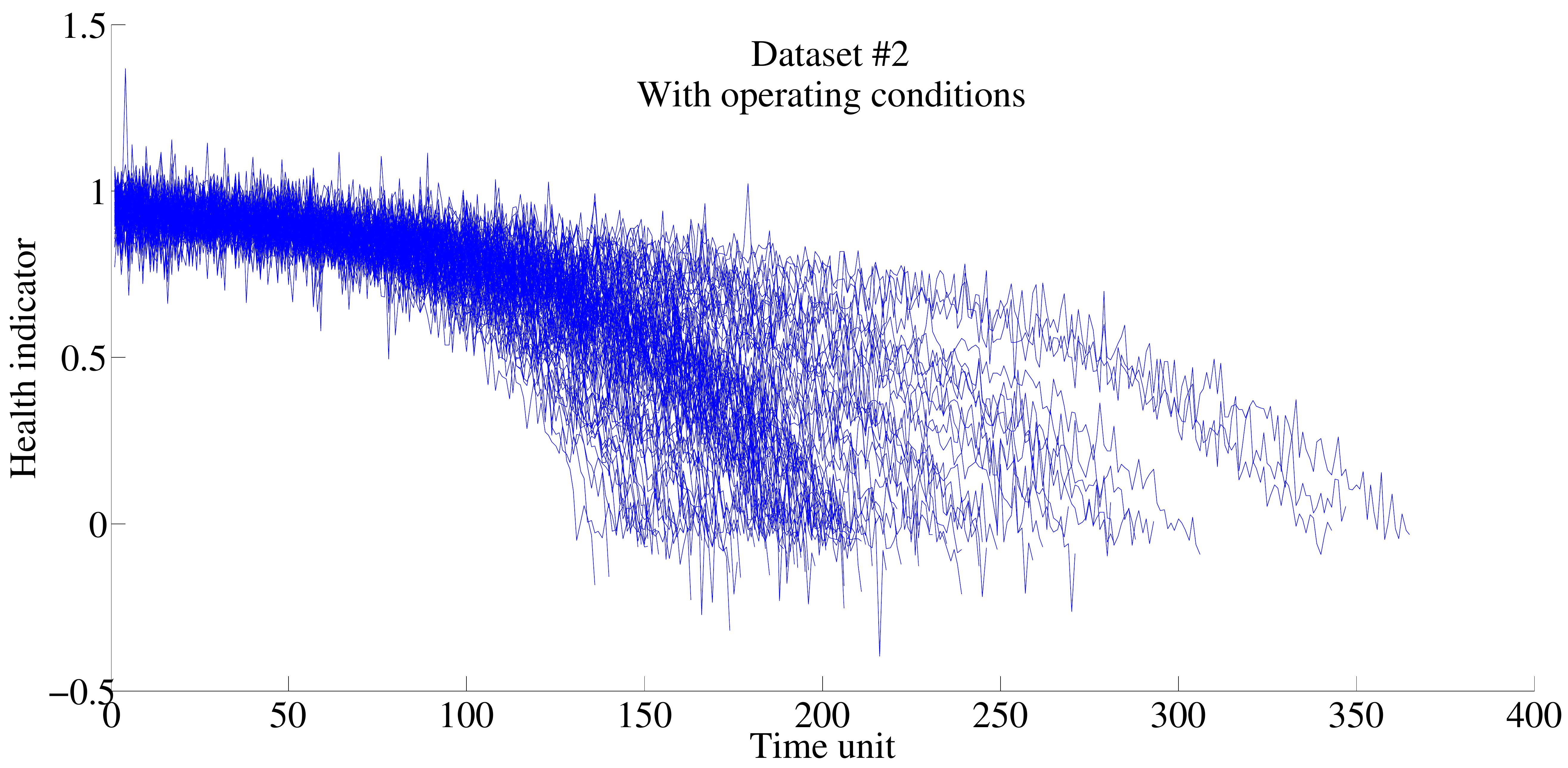}
                \caption{\#2 with operating conditions\label{2w}}
        \end{subfigure}        
        \begin{subfigure}[b]{0.45\textwidth}
                \includegraphics[width=0.95\textwidth]{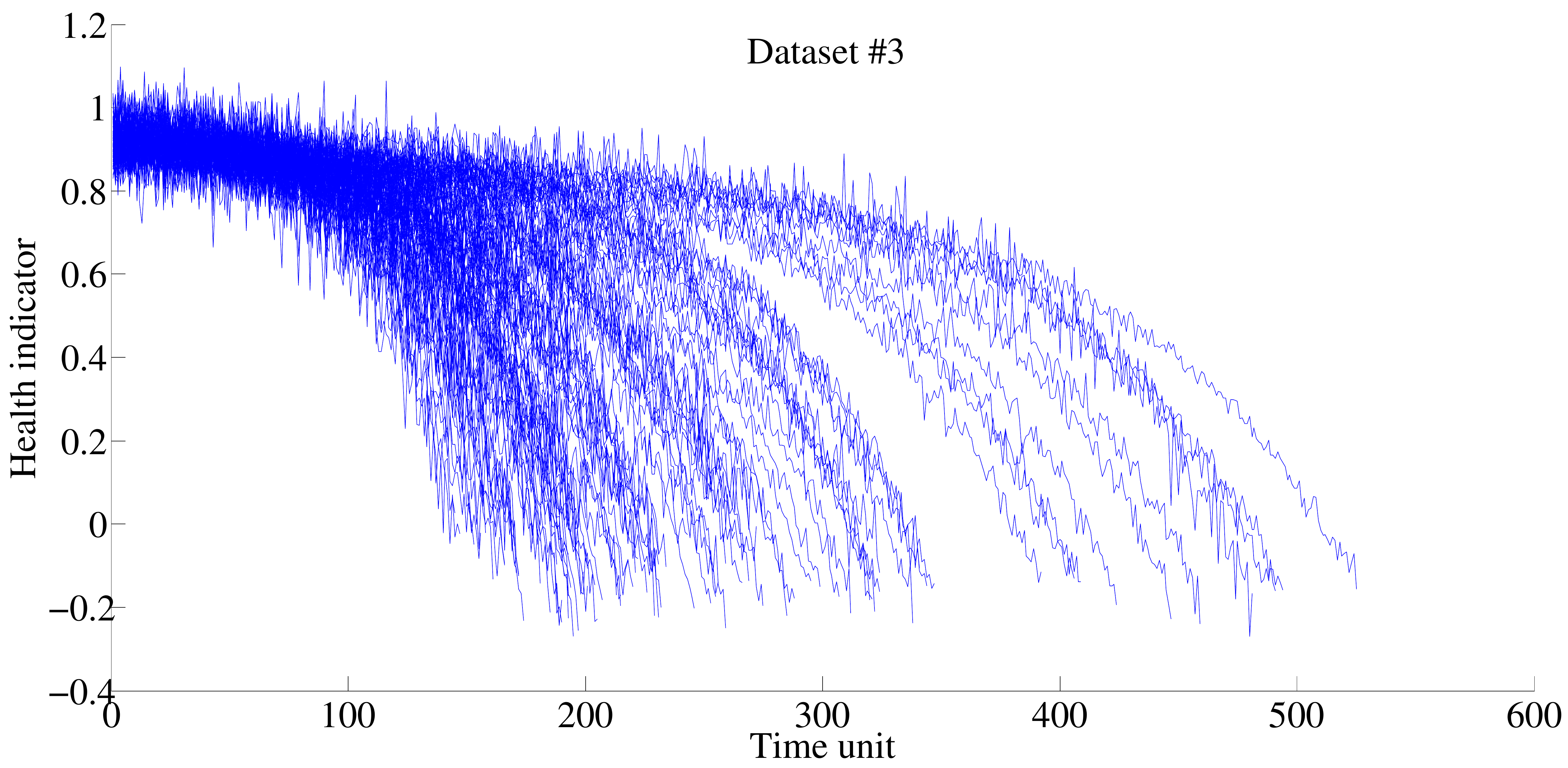}
                \caption{\#3}
        \end{subfigure}
        \centering
        \begin{subfigure}[b]{0.45\textwidth}
                \includegraphics[width=0.95\textwidth]{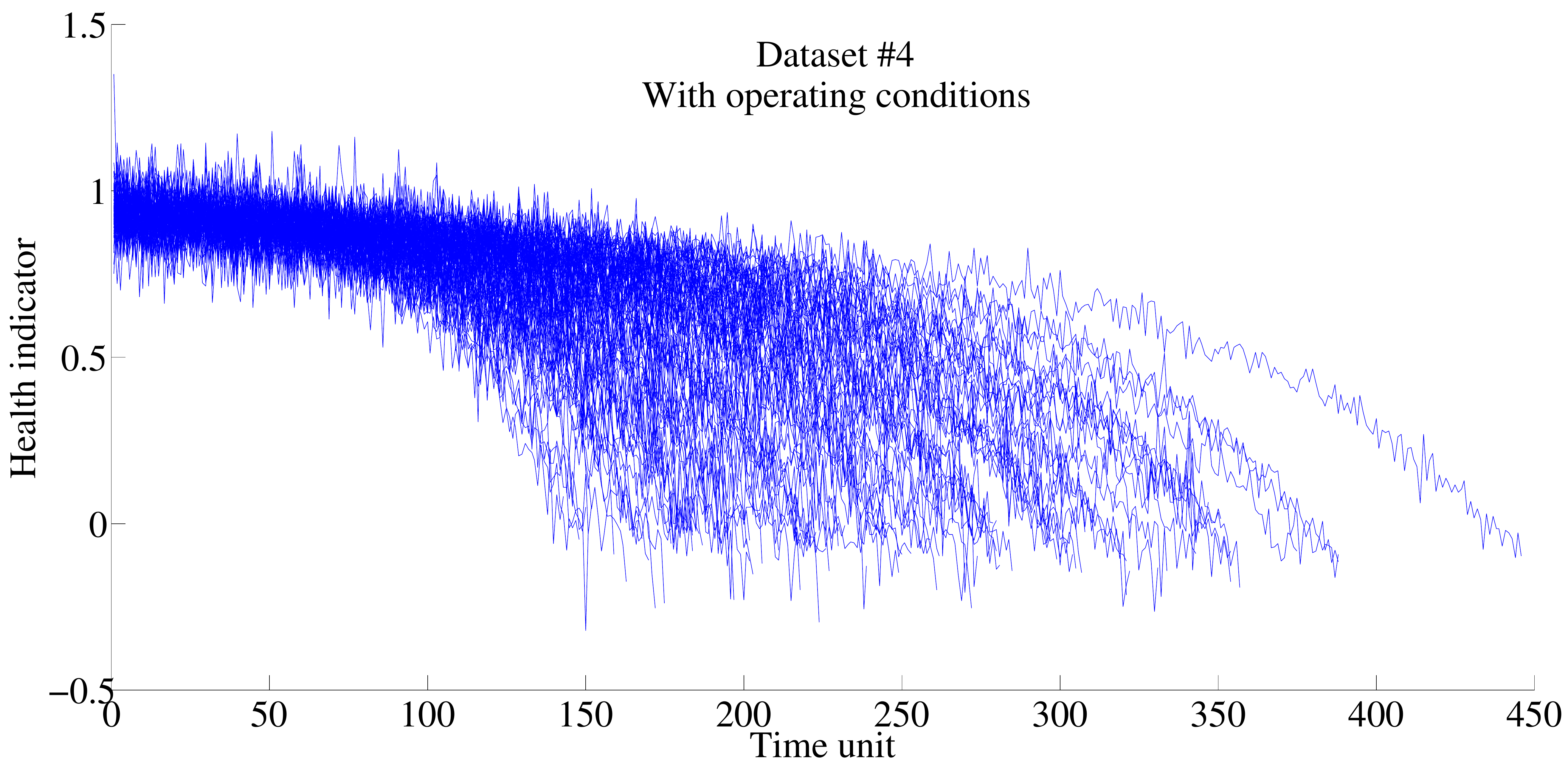}
                \caption{\#4 with operating conditions\label{4w}}
        \end{subfigure}      
        \caption{Evolution of the health indices for all engines in the four datasets.}
        \label{fig:errorplotCL2}
\end{figure*}

Figure~\ref{fig:errorplotCL2} illustrates the health indicators (computed as suggested \cite{Ramasso14}
and inspired from \cite{Wang10phd}) for all training data in each dataset. Dataset $\#1$ is made of $100$ training instances with 
an unique operating condition (OC) and unique fault mode, dataset $\#2$ with $260$ training instances, 
six OCs and one fault mode, dataset $\#3$ with $100$ training instances, one OC and two fault modes, 
and dataset $\#4$ with $248$ training instances made of six OCs and two fault modes.  

In the present paper, the $100$ training instances of the turbofan dataset $\#1$
were used for learning and 
the first $15$ testing for testing and comparison is made with 
RULCLIPPER algorithm \cite{Ramasso14} (available at the following
web page:
{\small \url{https://github.com/emmanuelramasso}}, 
with the ensemble-approach described in Table 7 of the latter publication).  
It is also compared to the Summation Wavelet Extreme Learning Machine (SWELM) 
algorithm proposed in \cite{javedIECON}.

\subsection{Prior information on the latent variables of the ARPHMM}

\begin{figure}[ht]
 \centering 
{\includegraphics[width=0.95\columnwidth]{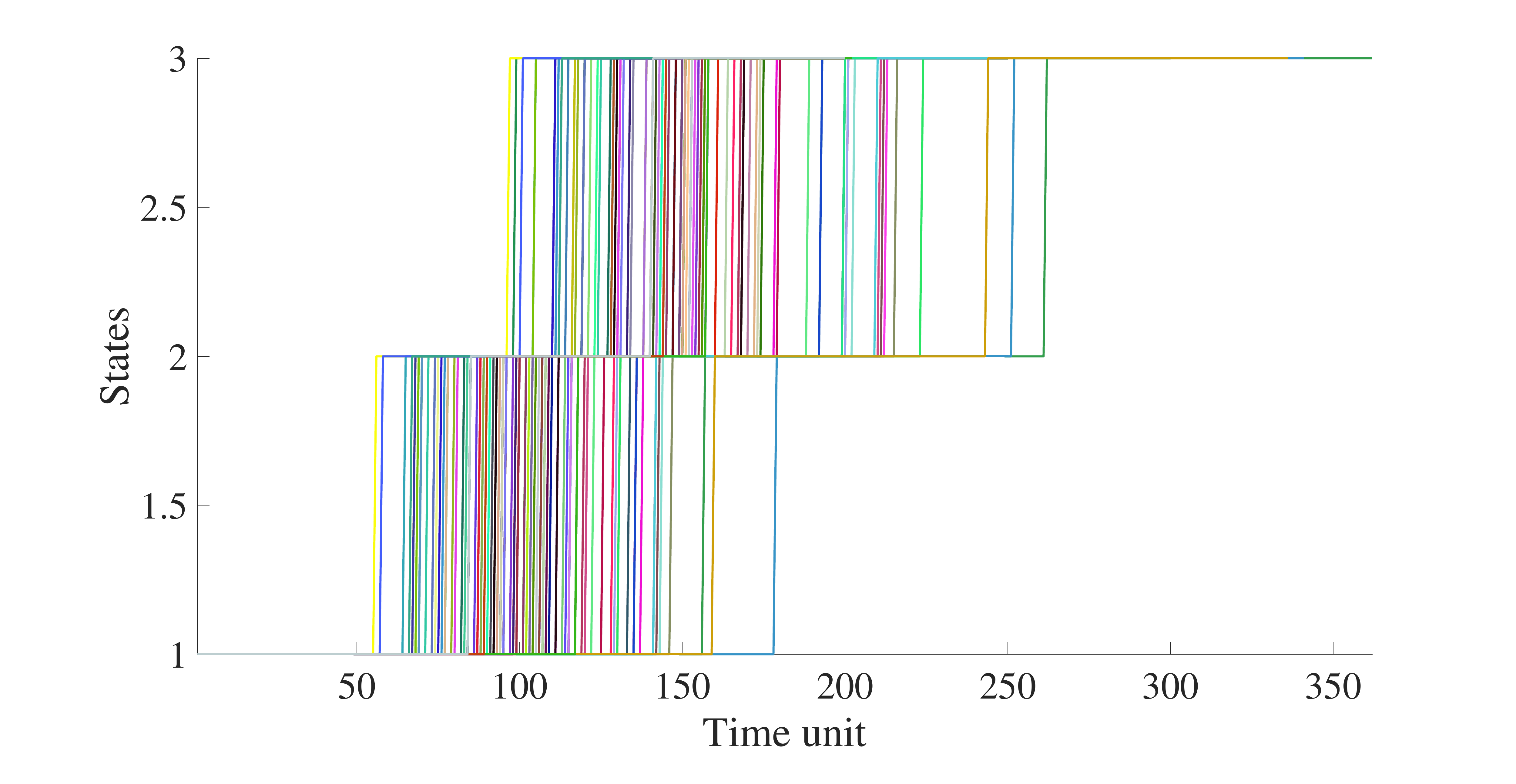}} 
\caption{Evolution of the finite levels of degradation. \label{fig:dsfef}}
\end{figure}

The ARPHMMs were trained using the health indicators shown in Figure~\ref{fig:hida1} 
(and available at the aforementionned web page) as inputs and using the associated RUL as output.
The real internal states of the turbofan are unknown but we insert some 
prior about some macroscopic latent variables by considering 
artificial finite degradation levels described in \cite{ramassoTR12016} (also available 
on the web page). Those levels, estimated by this method, are illustrated in Figure \ref{fig:dsfef}
for the $100$ training data.

The number of states for each training instance was thus equal to the number of states 
provided by the artificial degradation levels (3), 
and the number of regressors was equal to $7$ for all instances (about a quarter 
of the shorter testing instance). Note that the optimization of those parameters requires to 
develop some objective criteria with respect to the prior on latent variables which will 
be performed in future work.

The comparison between the estimated RUL and the ground truth provided at NASA 
PCOE was made using the timeliness function
as described in \cite{saxena2008damage} and with the mean average percentage error (MAPE)
which is possible since the RULs are available for dataset $\#1$.

\begin{table*}[htpb]
\caption{Results of ARPHMM and comparison. The timeliness S and the MAPE should be minimized. \label{tab:res}} 
\begin{tabular}{|c|c|c||c|c|c||c|c|c||c|c|c|}
\hline
Testing & Critical & True & \multicolumn{3}{|c|}{ARPHMM} & \multicolumn{3}{|c|}{SWELM} & \multicolumn{3}{|c|}{RULCLIPPER} \\
instance & time & RUL & \multicolumn{3}{|c|}{proposed} & \multicolumn{3}{|c|}{\cite{javedIECON}} & \multicolumn{3}{|c|}{\cite{Ramasso14}} \\
\cline{4-12}
  & & &   $\hat{\textrm{RUL}}$ & \;\;\;S\;\;\; & MAE & $\hat{\textrm{RUL}}$ & \;\;\;S\;\;\; & MAE & $\hat{\textrm{RUL}}$ & \;\;\;S\;\;\; & MAE \\
\hline
1 & 31 & 112 & 115.61 & 0.435 & 3.222 & 112 & 0 & 0 & 121.95 & 1.705 & 8.884 \\
\hline
2 & 49 & 98 & 107.00 & 1.460 & 9.184 & 54 & 28.507 & 44.898 & 108.03 & 1.727 & 10.236 \\
\hline
3 & 126 & 69 & 78.09 & 1.481 & 13.170 & 68 & 0.08 & 1.449 & 57.960 & 1.338 & 15.998\\
\hline
4 & 106 & 82 & 79.85 & 0.180 & 2.619 & 80 & 0.166 & 2.439 & 76.832 & 0.488 & 6.302 \\
\hline
5 & 98 & 91 & 87.85 & 0.274 & 3.459 & 100 & 1.459 & 9.890 & 78.119 & 1.693 & 14.155 \\
\hline
6 & 105 & 93 & 80.85 & 1.546 & 13.062 & 108 & 3.482 & 16.129 & 102.173 & 1.502 & 9.863 \\
\hline
7 & 160 & 91 & 78.17 & 1.682 & 14.095 & 114 & 8.974 & 25.274 & 97.015 & 0.824 & 6.605\\
\hline
8 & 166 & 95 & 75.226 & 3.577 & 20.815 & 102 & 1.014 & 7.368 & 95.886 & 0.093 & 0.933 \\
\hline
9 & 55 & 111 & 108.58 & 0.205 & 2.180 & 105 & 0.587 & 5.405 & 117.586 & 0.932 & 5.933 \\
\hline
10 & 192 & 96 & 73.80 & 4.514 & 23.121 & 68 & 7.618 & 29.167 & 87.219 & 0.965 & 9.146 \\
\hline
11 & 83 & 97 & 95.11 & 0.156 & 1.945 & 67 & 9.051 & 30.928 & 104.005 & 1.015 & 7.222 \\
\hline
12 & 217 & 124 & 85.93 & 17.692 & 30.700 & 131 & 1.014 & 5.645 & 100.939 & 4.894 & 18.598 \\
\hline
13 & 195 & 95 & 83.42 & 1.438 & 12.192 & 92 & 0.259 & 3.158 & 84.444 &  1.252 & 11.112 \\
\hline
14 & 46 & 107 & 112.62 & 0.755 & 5.255 & 81 & 6.389 & 24.299 & 119.931 & 2.644 & 12.085 \\
\hline
15 & 76 & 83 & 89.56 & 0.928 & 7.910 & 106 & 8.974 & 27.711 & 94.711 & 2.225 & 14.109 \\
\hline
 \multicolumn{3}{|c||}{Overall performance} & - & \textbf{36.32} & \textbf{10.86\%} & - & \textbf{77.57} & \textbf{15.58\%} & - & \textbf{23.30} & \textbf{10.08\%} \\
 \hline
 \multicolumn{3}{c}{ } &  \multicolumn{6}{|c|}{ } & \multicolumn{3}{c}{ } \\
 \multicolumn{3}{c}{ } &  \multicolumn{6}{|c|}{\bf FUSION} & \multicolumn{3}{c}{ } \\
 \multicolumn{3}{c}{ } &  \multicolumn{6}{|c|}{ } & \multicolumn{3}{c}{ } \\
 \cline{4-9}
 \multicolumn{3}{c}{ } & \multicolumn{3}{|c}{ } & 
 \multicolumn{3}{c|}{ }  & \multicolumn{3}{c}{ } \\
 \multicolumn{3}{c}{ } & \multicolumn{3}{|c}{\underline{S=\textbf{21.8}}} & 
 \multicolumn{3}{c|}{\underline{MAPE=\textbf{9.26\%}}}  & \multicolumn{3}{c}{ } \\
 \multicolumn{3}{c}{ } & \multicolumn{3}{|c}{ } & 
 \multicolumn{3}{c|}{ }  & \multicolumn{3}{c}{ } \\
 \cline{4-9}

 \end{tabular}
\end{table*}

Results are gathered in Table~\ref{tab:res}. Compared to RULCLIPPER, which 
provided the best results on dataset $\#1$ with few parameters \cite{RamassoSaxena2014}, 
the ARPHMM provides quite similar results in term of MAPE but less in term of timeliness. 
This result is encouraging since no optimization was performed for the number of states nor 
the number of regressors. 
The accuracy is around $80\%$ when considering the interval $[-13, 10]$ around the ground truth
with a false positive rate around $67\%$ corresponding to early predictions. 

ARPHMM provides better results on average compared to SWELM on those samples.
However, one can observe that the fusion 
of elements using a simple average of RUL estimates allows to get better results than RULCLIPPER
for those particular 15 instances. 

The comparison can not be generalized when compared to RULCLIPPER since it 
has demonstrated robustness with few parameters on all turbofan datasets
(with OC and two fault modes) using full testing datasets as well as on the PHM data challenge. 
However, the results go in favor of developing ensemble approaches made 
of complementary and advanced prognostics algorithms. This is all the more true than 
algorithms' parameterizations play an important role on the robustness which may be criticial 
for in-service use.

\subsection{Behavior of the proposed model}

\begin{figure*}
        \centering
        \begin{subfigure}[b]{0.49\textwidth}
                \includegraphics[width=\textwidth]{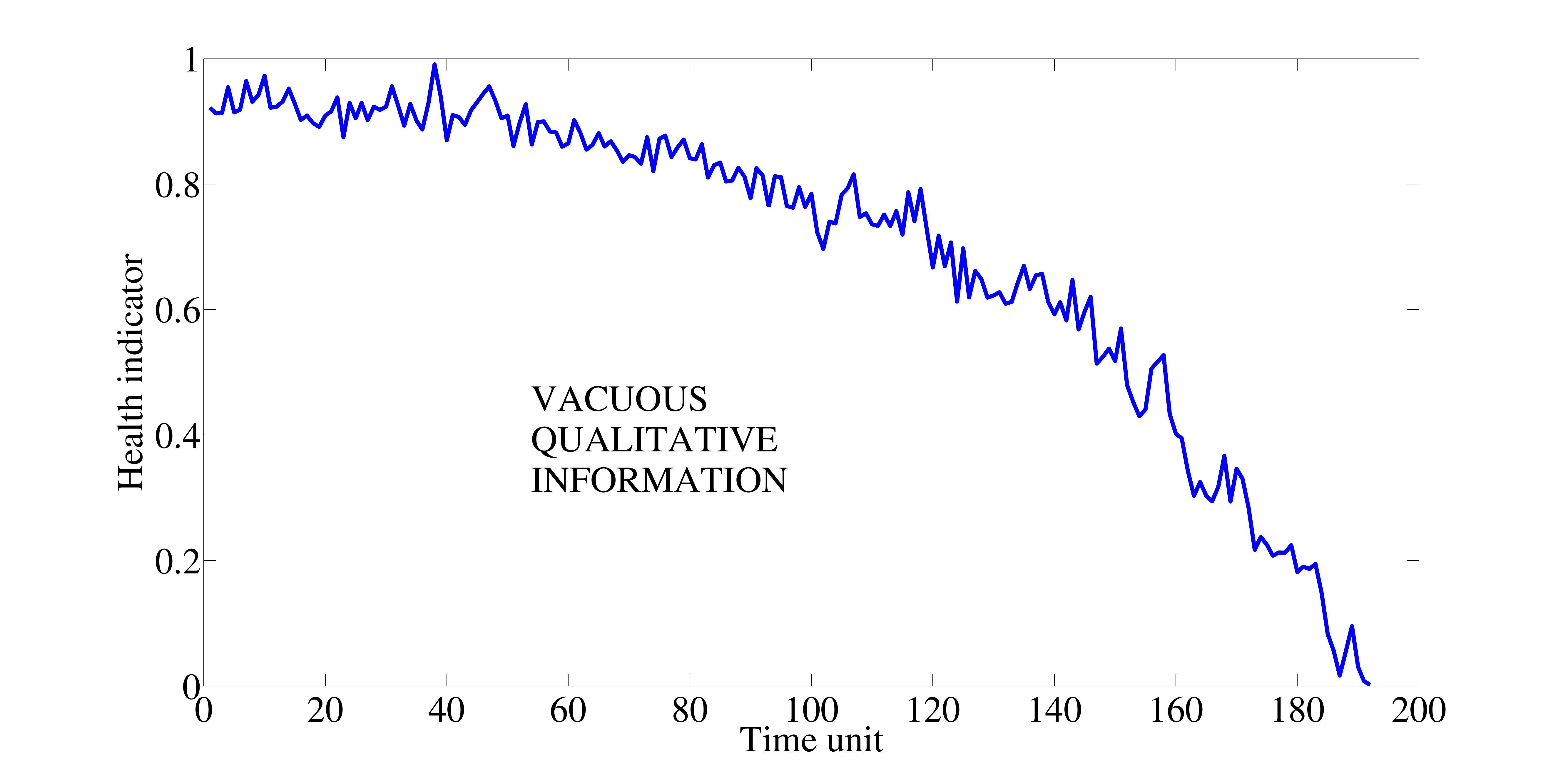}
                \caption{Vacuous knowledge (only the instance is provided).\label{fig:hida1sfsefse}}
        \end{subfigure}
        \centering
        \begin{subfigure}[b]{0.49\textwidth}
                \includegraphics[width=\textwidth]{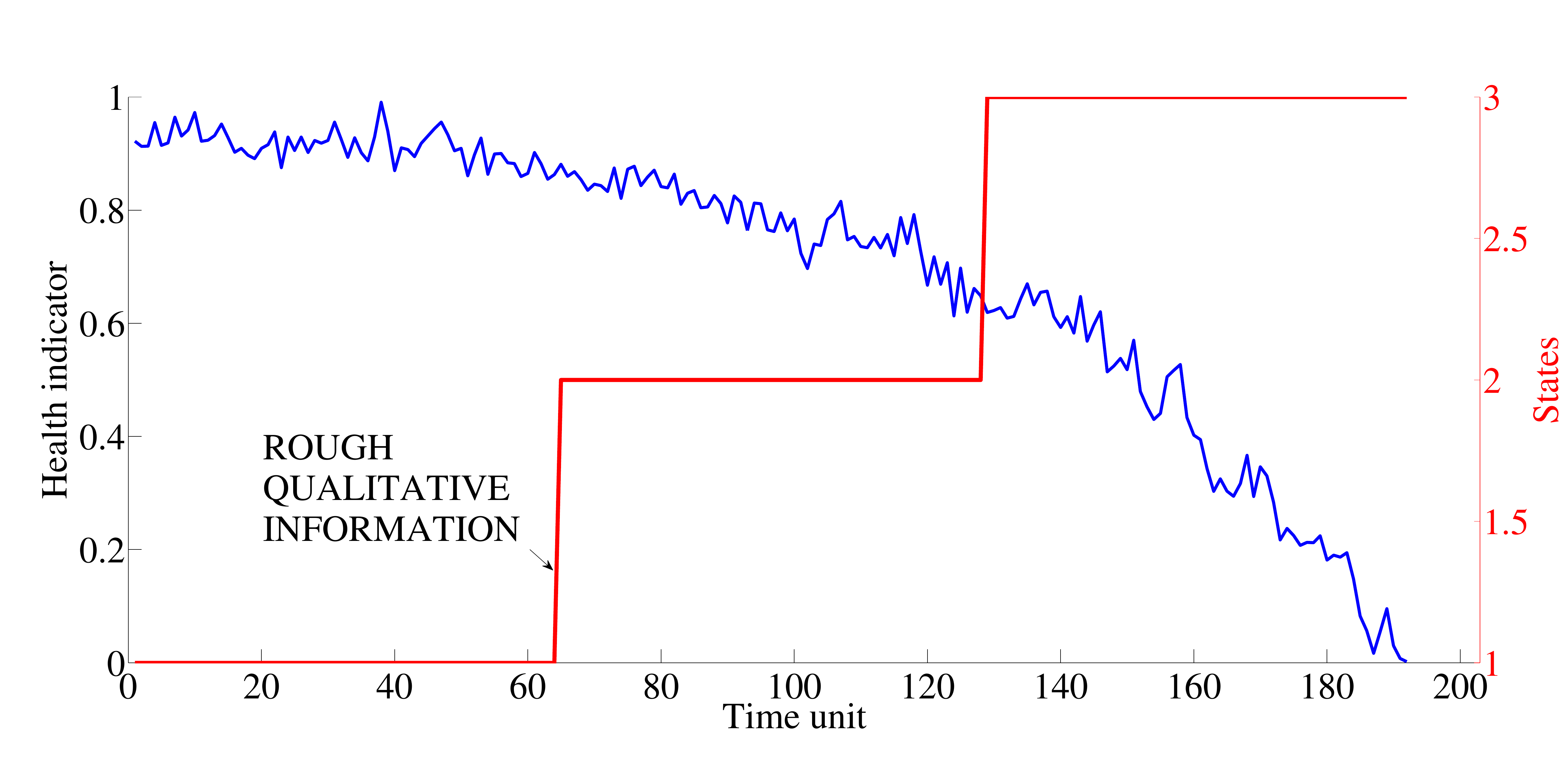}
                \caption{Rough knowledge on states.\label{2wlksljflkzjf}}
        \end{subfigure}        
        \begin{subfigure}[b]{0.49\textwidth}
                \includegraphics[width=\textwidth]{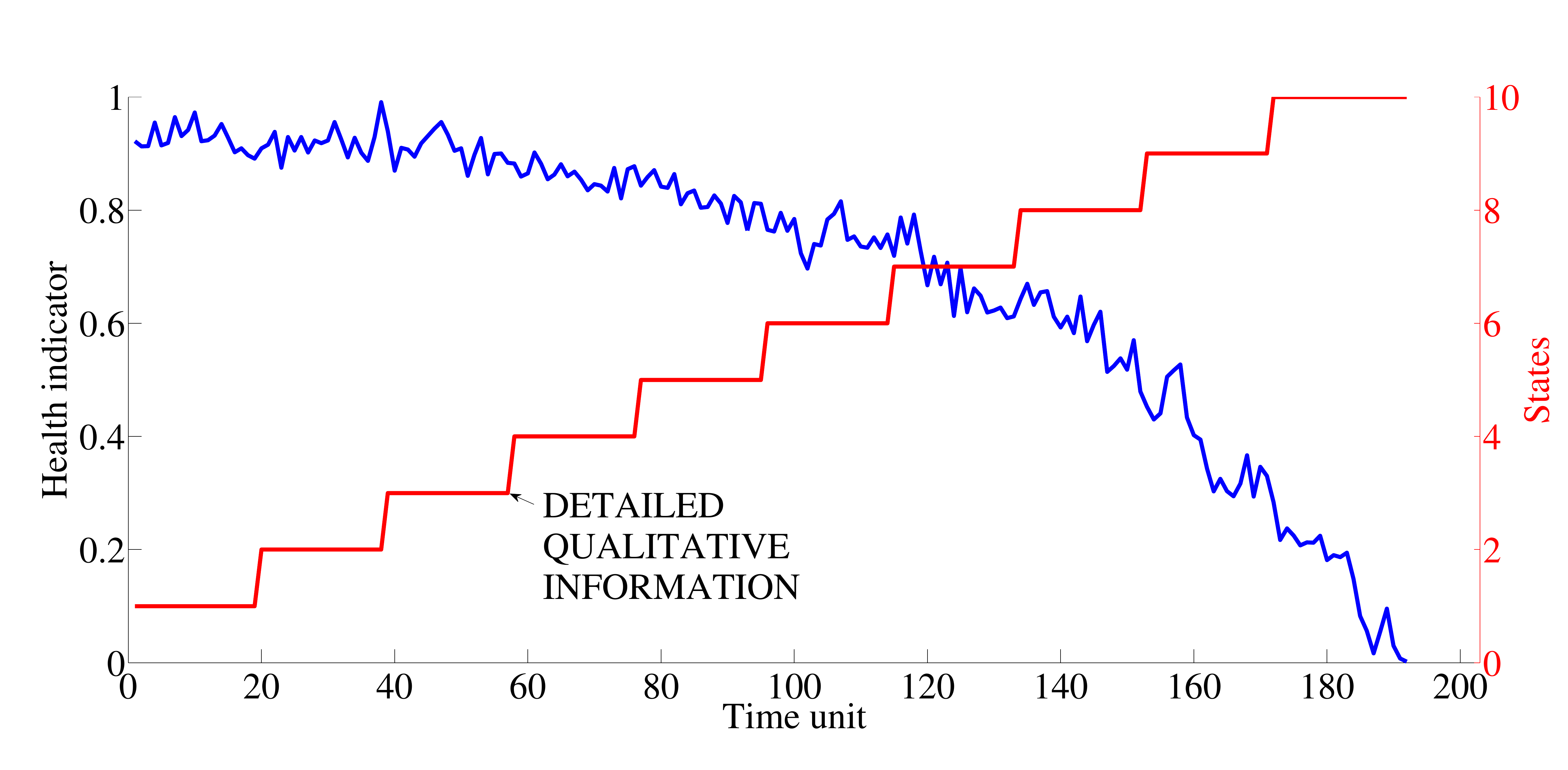}
                \caption{Detailed knowledge on states. \label{fergege}}
        \end{subfigure}
        \caption{Training instance 1, and different qualitative degradation levels
        used for training.}
        \label{fig:lkwjsclksjclkzejl}
\end{figure*}

Figure \ref{fig:lkwjsclksjclkzejl} illustrates the use of different hidden structure for learning 
the evolution of the first training instance in the dataset: K=3 states without (Fig.~\ref{fig:hida1sfsefse})
and with prior (Fig. \ref{2wlksljflkzjf}), and K=10 states (Fig.~\ref{fergege}). The training instance and the 
states are then used to learn one model, which is applied on the fourth training instance 
of the dataset (used as a testing instance). 

\begin{figure*}
        \centering
        \begin{subfigure}[b]{0.45\textwidth}
                \includegraphics[width=\textwidth]{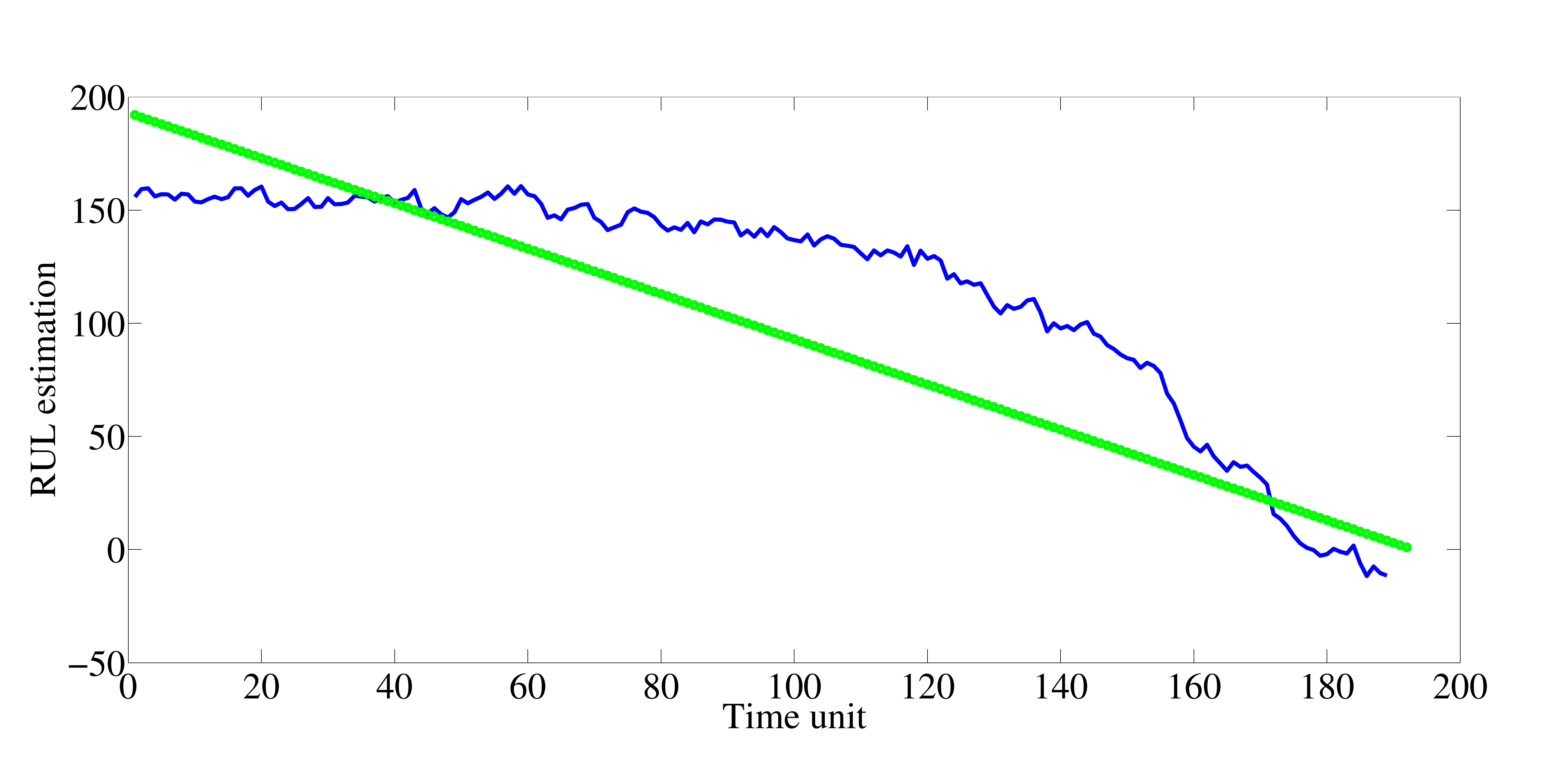}
                \caption{RUL estimation in the vacuous case.\label{rgqfzefzefzeryz}}
        \end{subfigure}
        \centering
        \begin{subfigure}[b]{0.45\textwidth}
                \includegraphics[width=\textwidth]{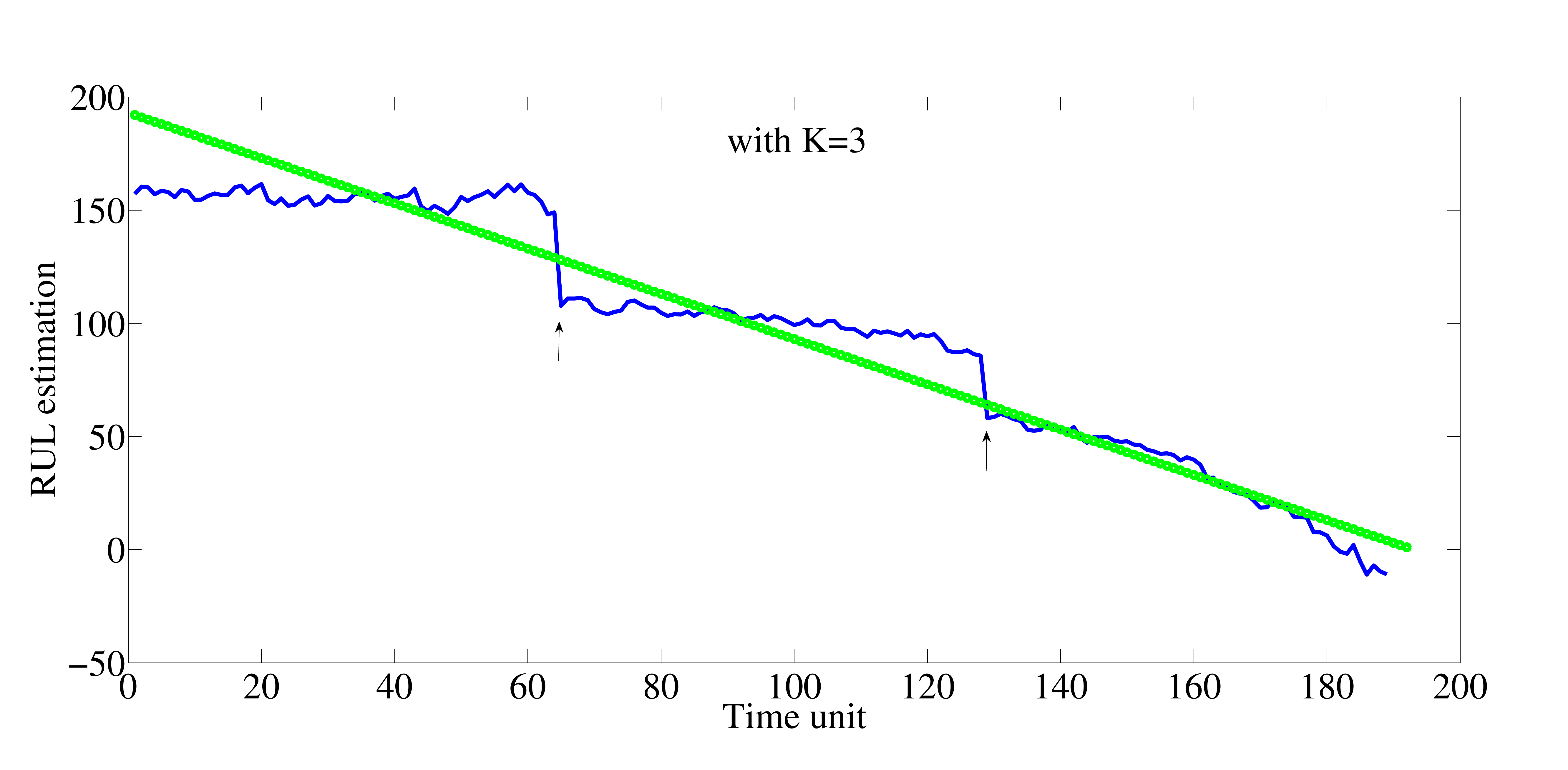}
                \caption{RUL estimation using 3 states. \label{rgzeryz}}
        \end{subfigure}        
        \begin{subfigure}[b]{0.45\textwidth}
                \includegraphics[width=\textwidth]{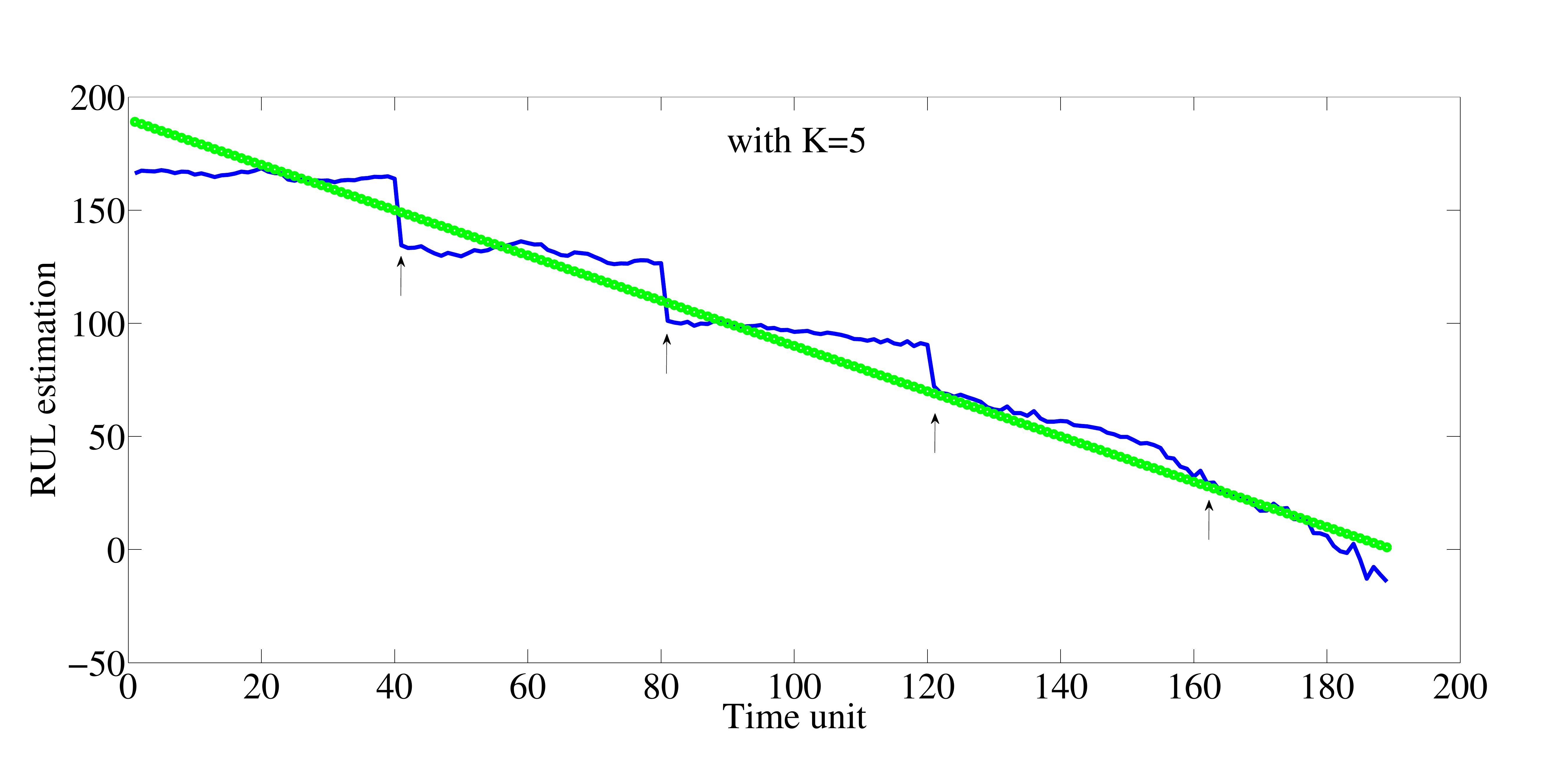}
                \caption{RUL estimation using 5 states.\label{egergerge}}
        \end{subfigure}
        \begin{subfigure}[b]{0.45\textwidth}
                \includegraphics[width=\textwidth]{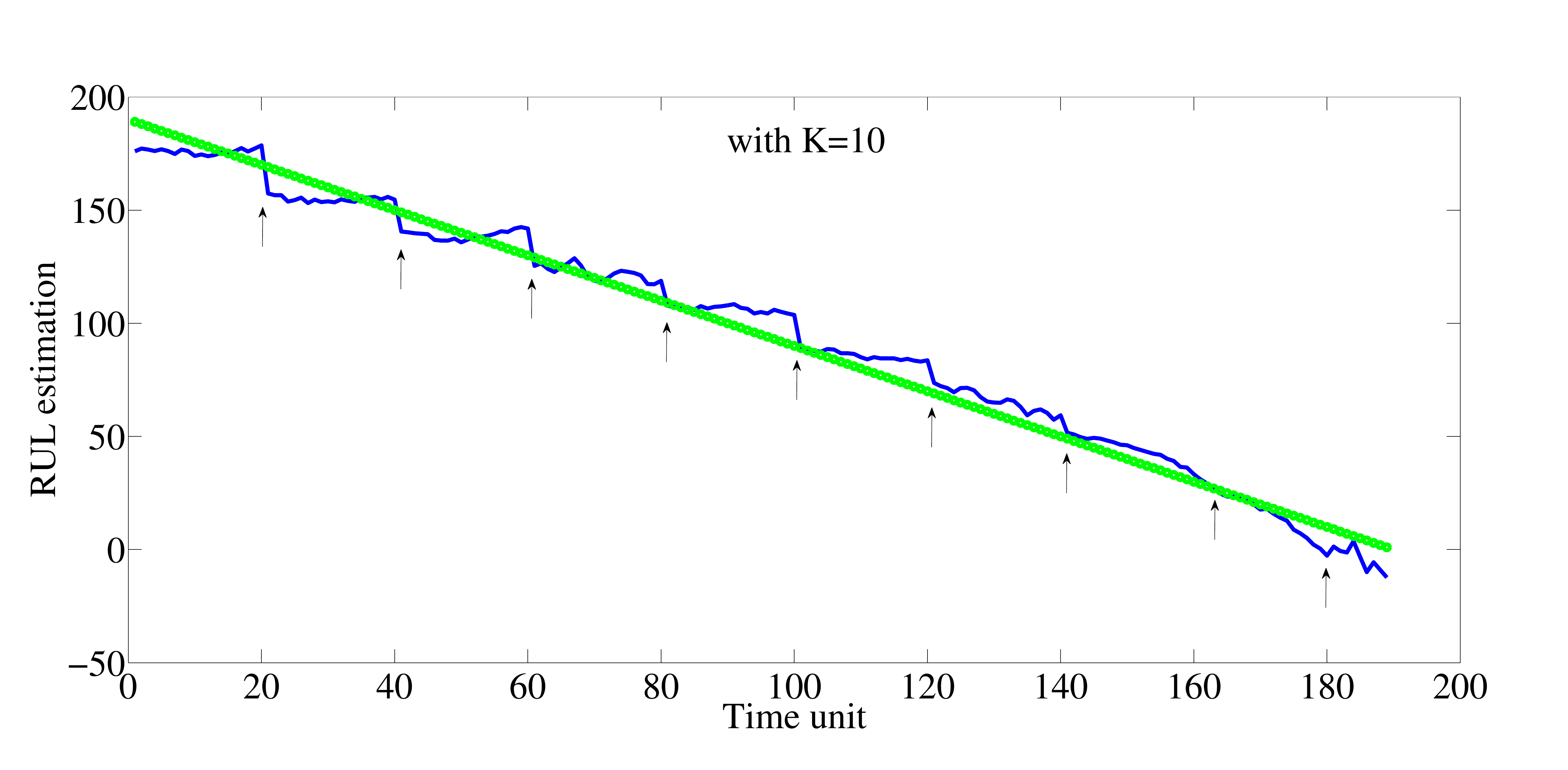}
                \caption{RUL estimation using 10 states.\label{seghergdfrd}}
        \end{subfigure}
        \caption{RUL estimation with different configurations on the hidden structure.}
        \label{fig:lkwjsclksjclkzsmlefkzmelkejl}
\end{figure*}

Figure \ref{fig:lkwjsclksjclkzsmlefkzmelkejl} is the application of this model for direct RUL estimation 
at each time step and for the testing instance. It can be observed that the model seems to 
provide better results on this new instance when more states are added. 

\begin{figure*}[ht]
 \centering 
  \centering
        \begin{subfigure}[b]{0.49\textwidth}
                {\includegraphics[width=1\textwidth]{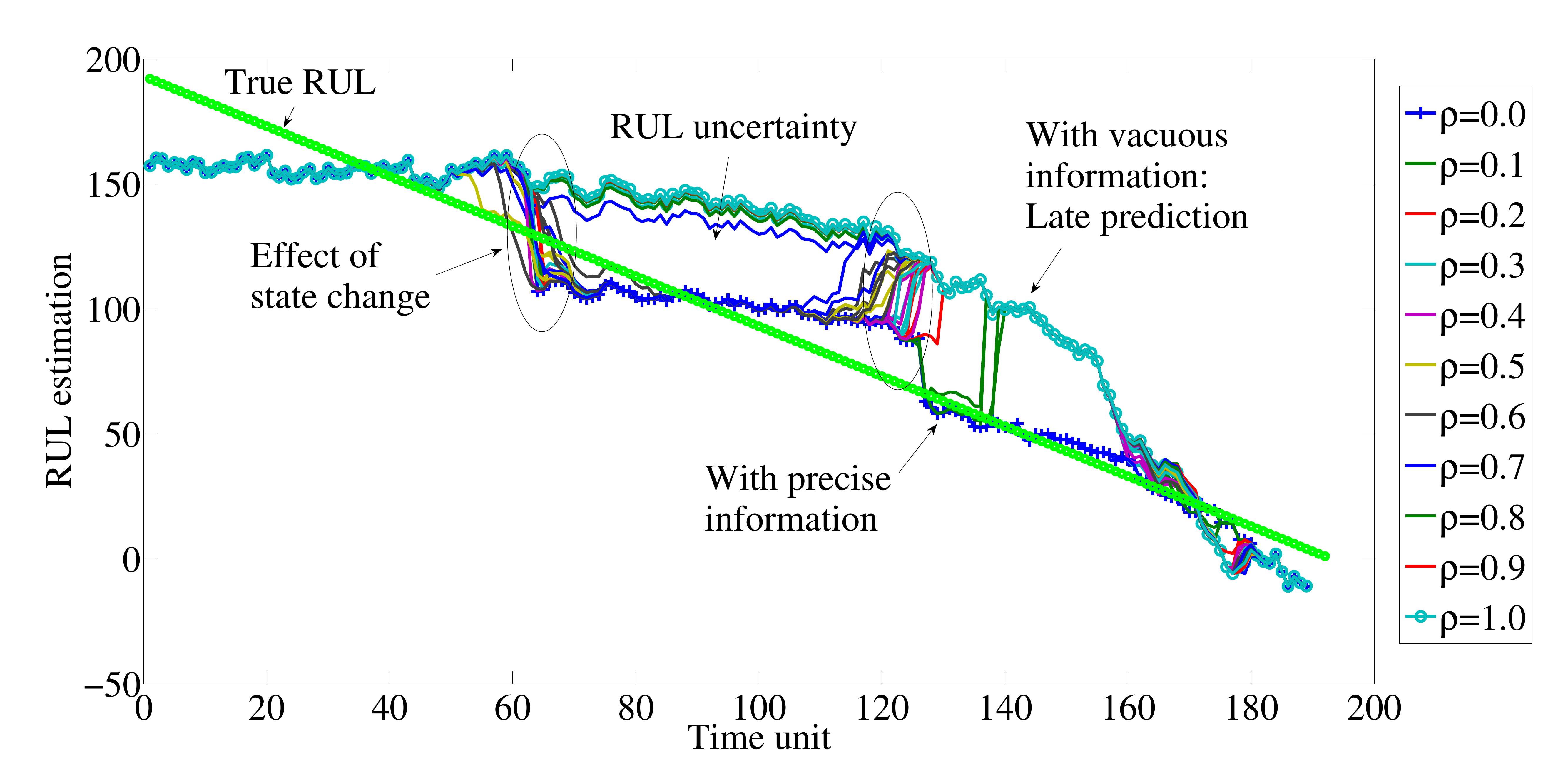}} 
                \caption{K=3.\label{rgqfzefzefzeryz}}
        \end{subfigure} 
        \centering
        \begin{subfigure}[b]{0.49\textwidth}
                \includegraphics[width=1\textwidth]{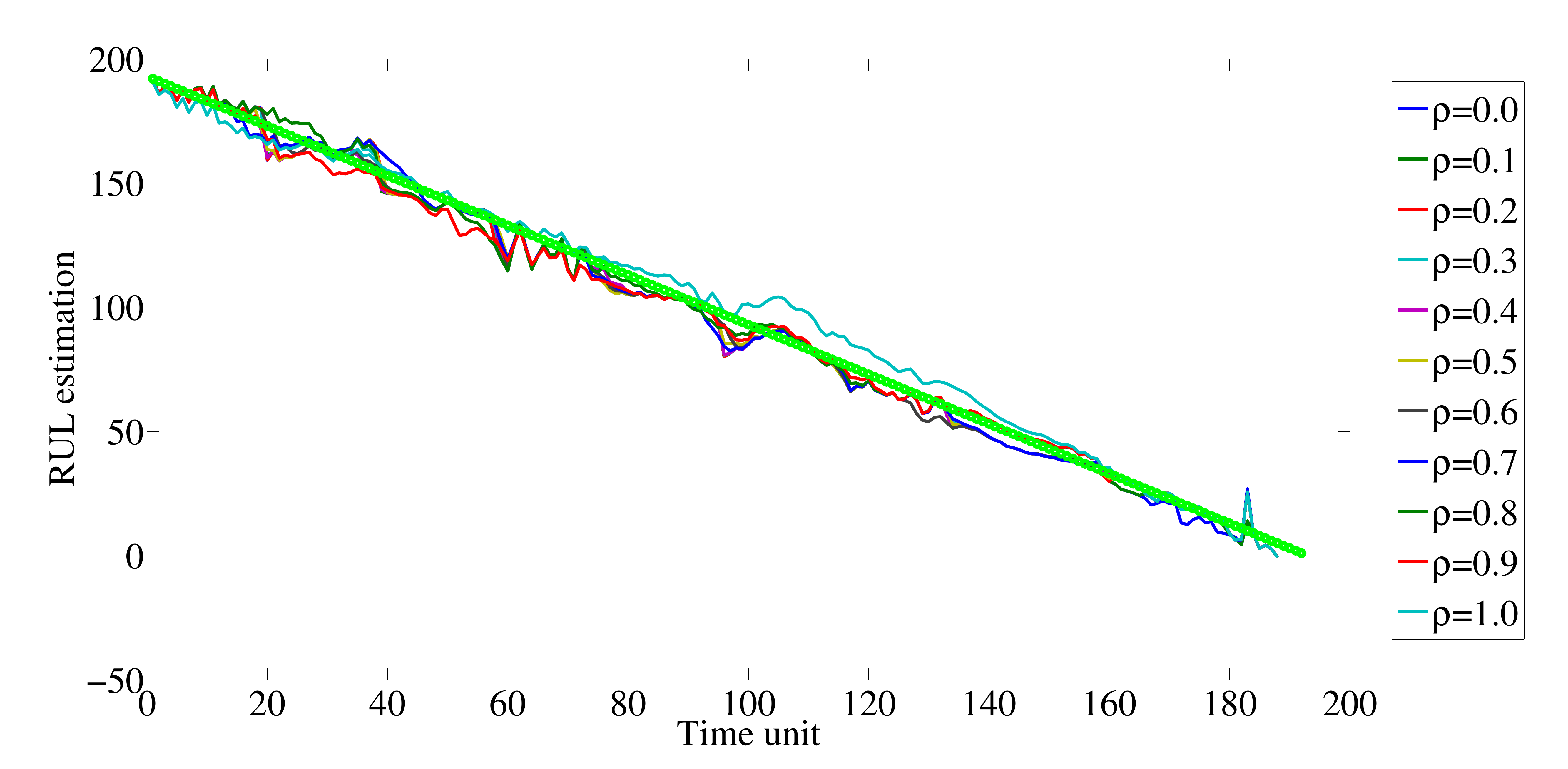}
                \caption{K=10. \label{rgzeryz}}
        \end{subfigure}              
\caption{RUL estimation as a function of the quality of prior ($\rho$) and states \label{fig:dsfefmkwsjlksj}}
\end{figure*}

Finally, Figure \ref{fig:dsfefmkwsjlksj} illustrates the impact of the quality of the prior. 
This case may correspond to a situation where new but partial knowledge is available and has 
to be integrated during prognostics (for instance information on the operating conditions). 
The quality is varied using a random sampling of the uncertainty on states as proposed 
in \cite{come2009learning} (code available at the aforementionned web page). The sampling process is governed 
by a parameter $\rho \in [0,1]$ such that $\rho=0$ corresponds to the supervised
(full quality) case, $\rho=1$ to the unsupervised case (no prior), and intermediary
values correspond to noisy prior, all the more noisy than $\rho$ increases.

It can be observed that for $K=3$, the RUL estimation is highly dependent on the prior.
Besides, the uncertainty can be quantified with different values of $\rho$. Uncertainty is much higher 
is the elbow part of the degradation, while it is quite low for the beginning (since all 
data have similar evolution) and for the end (when converging to the solution). 
For $K=10$, the RUL estimation is more accurate and does not depend on the prior.

Those figures also remind prognostics approaches based on multi-modelling such as \cite{Serir11} and
\cite{serir2012evidential}. As for multimodels, the hidden structure in an ARPHMM allows to evaluate 
the active feature space, while, for each state, the evolution of the time-series is approximated. 
However, in addition to quantifying the uncertainty on states for each new measurement, 
 the ARPHMM additionally quantifies the likelihood associated to a model which makes model 
 selection possible.

\section{Conclusion and perspectives} 

In this paper, we investigate the use of uncertain prior on the latent structure of dynamic Bayesian
network for prognostics, and in particular an autoregressive partially hidden Markov model. 
More experiments are needed to validate the approach but results obtained on some instances of CMAPSS
datasets are encouraging when compared to other approaches from the literature. 
This model is being improved to include uncertain future operating conditions on the latent structure.

\section*{Acknowledgment}

This work has been carried out in the framework of the Laboratory of
Excellence ACTION through the program ``Investments for the future'' managed
by the National Agency for Research (references ANR-11-LABX-01-01). 
The authors are grateful to the R\'egion Franche-Comt\'e and ``Bpifrance 
financement'' supporting the SMART COMPOSITES Project in the framework of FRI2.
The authors also thank the anonymous reviewers who provided useful comments and 
advices to improve the paper.

%
%
%
%

\bibliographystyle{unsrt}  
\bibliography{articledhmmbib}  

\end{document}